\documentclass{article}

\usepackage{arxiv}

\usepackage[utf8]{inputenc} 
\usepackage[T1]{fontenc}    
\usepackage{hyperref}       
\usepackage{url}            
\usepackage{booktabs}       
\usepackage{amsfonts}       
\usepackage{nicefrac}       
\usepackage{microtype}      
\usepackage{lipsum}
\usepackage{graphicx}
\usepackage{amsmath}
\graphicspath{ {./images/} }

\title{Enhancing Multi-Modal Video Sentiment Classification Through Semi-Supervised Clustering}

\author{
 Mehrshad Saadatinia$^*$  \\
  Viterbi School of Engineering\\
  University of Southern California\\
  Los Angeles, CA \\
  \texttt{saadatin@usc.edu} \\
   \And
 Minoo Ahmadi$^*$  \\
  Viterbi School of Engineering\\
  University of Southern California\\
  Los Angeles, CA \\
  \texttt{minooahm@usc.edu} \\
  \And
 Armin Abdollahi$^*$  \\
  Viterbi School of Engineering\\
  University of Southern California\\
  Los Angeles, CA \\
  \texttt{arminabd@usc.edu} \\
  \vspace{0.8em} 
  \small{$^*$These authors contributed equally to this work.}
}

\begin{document}
\maketitle
\begin{abstract}
\textit{Understanding emotions in videos is a challenging task. However, videos contain several modalities which make them a rich source of data for machine learning and deep learning tasks. In this work, we aim to improve video sentiment classification by focusing on two key aspects: the video itself, the accompanying text, and the acoustic features. To address the limitations of relying on large labeled datasets, we are developing a method that utilizes clustering-based semi-supervised pre-training to extract meaningful representations from the data. This pre-training step identifies patterns in the video and text data, allowing the model to learn underlying structures and relationships without requiring extensive labeled information at the outset. Once these patterns are established, we fine-tune the system in a supervised manner to classify the sentiment expressed in videos. We believe that this multi-modal approach, combining clustering with supervised fine-tuning, will lead to more accurate and insightful sentiment classification, especially in cases where labeled data is limited.}
\end{abstract}

\textbf{Keywords:} Video sentiment classification, multimodal learning, clustering, pre-training, semi-supervised learning.


\section{Introduction}

In our study, we aim to tackle the challenge of enhancing video sentiment classification by addressing the complexities of analyzing visual expressions and linguistic and acoustic features together. Traditional approaches in this field rely predominantly on supervised learning, which demands large labeled datasets—a significant limitation given the high costs and effort involved in annotating nuanced emotional data in videos. Our approach circumvents this dependency by leveraging clustering-based pre-training to capture latent patterns in video and text data, enabling the model to learn meaningful emotional representations on the CMU-MOSI dataset \cite{zadeh2018multimodal}. This pre-training strategy supports a subsequent supervised fine-tuning phase, which further refines the model's ability to classify sentiments accurately. Through this approach, we aim to improve sentiment analysis in applications such as social media monitoring, customer feedback analysis, and content recommendation, where precise emotional understanding can drive better decision-making and engagement. Key challenges include capturing subtle cross-modal interactions and emotional cues without extensive labeled data, but our method has the potential to offer practical benefits for companies, researchers, and users who rely on robust sentiment analysis systems.

Current video sentiment classification models, such as the Multimodal Transformer, capture cross-modal interactions via attention mechanisms, effectively integrating visual and textual inputs. However, these models are limited by their dependence on large labeled datasets like CMU-MOSI, which are expensive and challenging to produce. Recently, self-supervised methods, such as Contrastive Multiview Coding (CMC) \cite{tian2020contrastive}, have been explored as alternatives for learning from multimodal data without explicit labels. Although promising, they often fall short of capturing the nuanced emotional cues required for sentiment-specific tasks, as they primarily focus on aligning data views rather than on emotional features. Our proposed clustering-based pre-training approach leverages multimodal deep clustering to derive meaningful representations by using a large subset of unlabeled data alongside a smaller labeled set that guides cluster formation toward capturing sentiment. This data-efficient strategy enables the model to identify underlying emotional structures across modalities. After pre-training, supervised fine-tuning further enhances the model’s accuracy. Additionally, we hypothesize that by using this clustering based pre-training instead of typical unsupervised pre-training \cite{hinton2006reducing, vincent2008extracting} we obtain superior results. 

Our method could be phrased as semi-supervised. Unsupervised learning trains a model using only unlabeled data, relying on inherent patterns to learn representations without guidance. In contrast, semi-supervised learning combines a small amount of labeled data with a larger volume of unlabeled data, using the labeled samples to guide the model while still benefiting from the unlabeled data. This approach is effective when labeled data is limited or costly.

By evaluating this approach against supervised baselines with metrics such as accuracy and F1 score on the CMU-MOSI dataset, we demonstrate its effectiveness in handling limited labeled data, thus advancing the field of video sentiment classification and addressing current limitations.

\section{Related Work}

The rapid proliferation of user-generated content on social media platforms, encompassing text, images, audio, and video, has propelled the field of Multimodal Sentiment Analysis (MSA) into a critical area of research. MSA aims to understand and interpret sentiment across diverse modalities, leveraging the rich contextual information inherent in multimodal data to enhance sentiment prediction accuracy. Video is one of the richest data sources, as it simultaneously includes all essential modalities: text, audio, and visual frames. Current video sentiment classification approaches predominantly rely on fully-supervised methods, typically comprising three phases: encoding each modality, merging the features into a joint space, and classifying using a deep classifier \cite{abdu2021multimodal, gandhi2023multimodal}. Selecting effective encoding networks for different modalities has been an active research area. Many studies have employed variations of RNN-based networks, such as LSTM and GRU \cite{huan2021video, 8885108}, as well as attention-based or transformer-based \cite{vaswani2017attention} approaches \cite{kim2020multi, xi2020multimodal}, or a combination of both \cite{du2022gated}.
The choice of fusion networks for obtaining the joint embedding space has also been a topic of extensive research with different approaches ranging from simple concatenation, to attention-based and skip-connection-based methods \cite{jiao2024comprehensive, abdu2021multimodal}. 
The most widely-used benchmark datasets for multimodal sentiment classification on videos are CMU-MOSEI and CMU-MOSI. We chose CMU-MOSI to evaluate our method, as it facilitates reliable comparison with other approaches. Previous studies have typically used binary accuracy and F1 score to assess performance on this dataset; thus, we adopt these same metrics for our evaluation.

Among recent studies on MSA, the best performance on CMU-MOSI has been achieved by two methods: MARNN with 84.31\% accuracy \cite{kim2020multi} and MMUU-BA with 83.31\% accuracy \cite{ghosal-etal-2018-contextual}. Both approaches use GRU networks alongside attention and self-attention mechanisms, employing concatenation for fusion. MARNN utilizes a Transformer block to extract specialized encodings from each modality, referred to as utterances. These encodings are then fed into a Bidirectional-GRU and concatenated for final classification. MMUU-BA also adopts a similar approach, using Bi-GRUs to extract initial encodings, followed by cross-attention between modalities before concatenating the encodings into a joint space for classification.

While outperforming other approaches and reaching above 80\% accuracy. These two methods have the drawbacks of high architectural complexity which we aim to address in our method.

\subsection{Leveraging Semi-Supervised Pretraining}

Most prior studies have achieved results in the 60\% to 80\% accuracy range, with the primary differences among them lying in the arrangement and fusion techniques employed in their architectures. However, there remains substantial room to explore non-architectural techniques that can improve performance without increasing architectural complexity. One of the most effective methods to enhance supervised machine learning tasks is through unsupervised and semi-supervised pretraining or transfer learning. A major challenge in MSA is the scarcity of labeled data, particularly in video sentiment analysis, where annotation can be both costly and time-consuming \cite{caschera2016sentiment, tomita2023tf, moreno2021class}. As a solution, unsupervised pretraining enables models to learn valuable representations from unlabeled data, reducing reliance on labeled data. This pretraining step allows the model to extract meaningful patterns from multimodal data, which are then fine-tuned for sentiment classification using available labels. This two-step process—first unsupervised (or semi-supervised), then supervised—helps overcome the limitations of labeled data while maximizing the utility of the rich multimodal content available in the dataset \cite{xie2016unsupervised}. While some past studies have explored transfer learning across modalities \cite{rajan2021cross}, a significant gap remains in leveraging knowledge transfer from other tasks to improve MSA results. Additionally, to the best of our knowledge, no studies have utilized clustering for pretraining and knowledge transfer in a classification task.

\subsection{Our Contributions} 
\begin{enumerate} 
\item We propose a simple yet robust architecture based on transformers for binary sentiment classification on the CMU-MOSI dataset, leveraging three modalities (text, audio, and video) with significantly fewer parameters compared to state-of-the-art methods.

\item We introduce a semi-supervised pretraining approach for deep clustering based on the Deep Embedded Clustering (DEC) framework proposed by Xie et al. \cite{xie2016unsupervised}. This pretraining facilitates knowledge transfer, leading to improved supervised classification performance and enhancing the baseline classification accuracy.

\item We modify the DEC architecture to support multimodal and semi-supervised training, enabling it to effectively process and integrate multiple modalities for better clustering and classification results.
\end{enumerate}

\section{Methodology}

Our methodology consists of two phases to enhance multimodal sentiment classification on the CMU-MOSI dataset after appropriate preprocessing. In the first phase, we implement a baseline classification network to perform initial sentiment classification on a smaller labeled subset of the dataset and evaluate the effectiveness of our architecture. This baseline is compared against several established methods, including state-of-the-art models, to assess its initial performance.

Building on this foundation, the second phase involves implementing a Deep Embedded Clustering (DEC)-based semi-supervised clustering network to improve feature representations. The DEC-based network is first pretrained on the entire dataset, including both labeled and unlabeled instances, to capture complex multimodal relationships. The pretrained network weights are then transferred to the classification task, serving as initialization for fine-tuning with labeled data. This semi-supervised pretraining enhances classification accuracy, and the results are compared against the baseline to quantify the impact of the DEC-based approach.

\subsection{Dataset and Preprocessing}
Our study utilizes the CMU-MOSI dataset, a multimodal dataset for sentiment analysis that includes audio, visual, and textual features.   

\begin{itemize}
    \item \textbf{Textual Features}: These are \textit{text embeddings} derived from the Global Vectors for Word Representation (GloVe) model. Each word in a video transcript is embedded into a high-dimensional vector space based on GloVe's pretrained embeddings. These embeddings capture semantic relationships in text, enabling effective textual representation for models.

    \item \textbf{Visual Features}: This modality provides \textit{visual features} related to facial expressions. It uses facial action coding and landmark detection to capture emotions through expressions. These features include measurements of facial muscle movements, head positions, and visual cues associated with affective states, helping capture non-verbal sentiment cues.

    \item \textbf{Audio Features}: This is the \textit{audio feature set} focused on vocal qualities. It includes pitch, glottal source parameters, formants, and spectral characteristics, which reflect vocal emotion. These features help capture the affective content in voice, such as intonation and tone, crucial for emotional context in spoken language.
\end{itemize}

The dataset contains 2183 instances. Each instance in the dataset is represented as a sequence with a varying sequence length but is consistent across modalities per sample. To accommodate these varying lengths, we normalize and pad (or truncate) the sequences to the same length of 30 and standardize input dimensions for training.
The preprocessing pipeline includes min-max scaling and z-score normalization of features across each modality to ensure uniform scaling and centered distributions, which are essential for the effective performance of the neural network. The labels in the original CMU-MOSI dataset are positive or negative integers. Negative numbers are treated as the negative sentiment label, while positive numbers are treated as the positive sentiment label for downstream binary classification. 

Additionally, to simulate the scenario of a small labeled set and a large unlabeled set, we removed the labels for 60\% of the dataset instances to create a larger unlabeled subset, using only the remaining 40\% for classification. However, the entire dataset was utilized for semi-supervised pretraining.

\subsection{Baseline Sentiment Classification}

For our baseline sentiment classification, we designed a multimodal neural network that processes three distinct modalities from the CMU-MOSI dataset: facial expressions, textual features, and acoustic features. Each modality is encoded using a dedicated Transformer-based encoder to capture its unique sequential dependencies and extract modality-specific features. The Transformer encoders include a positional encoding layer and two layers of Transformer blocks, each with 4 attention heads and 256-dimensional feed-forward networks. This architecture enables the encoders to effectively capture nuanced temporal patterns within each modality, resulting in robust modality-specific embeddings.

The model architecture, combines the mean-aggregated embeddings from each modality after encoding. These embeddings are concatenated to form a joint representation, which is passed through a fully connected layer, followed by batch normalization, ReLU activation, and dropout for regularization. This integrated multimodal representation is then processed through additional fully connected layers to progressively reduce the embedding dimension, ultimately outputting a single logit representing the sentiment score.

\begin{figure}[h]
    \centering
    \includegraphics[width=0.75\textwidth]{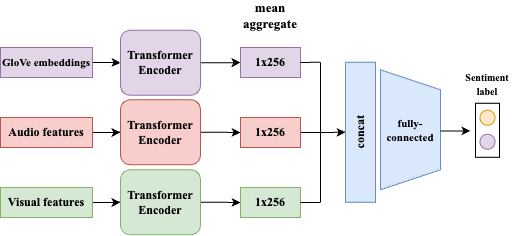}
    \caption{Network architecture schematics for the classification network}
    \label{fig:schema}
\end{figure}

This approach leverages the strengths of Transformer encoders for feature extraction and integrates multimodal information effectively. Figure \ref{fig:schema} illustrates the overall architecture of the network described in this section.

\subsection{Deep Embedded Multimodal Clustering}

\begin{figure}[h]
    \centering
    \includegraphics[width=\textwidth]{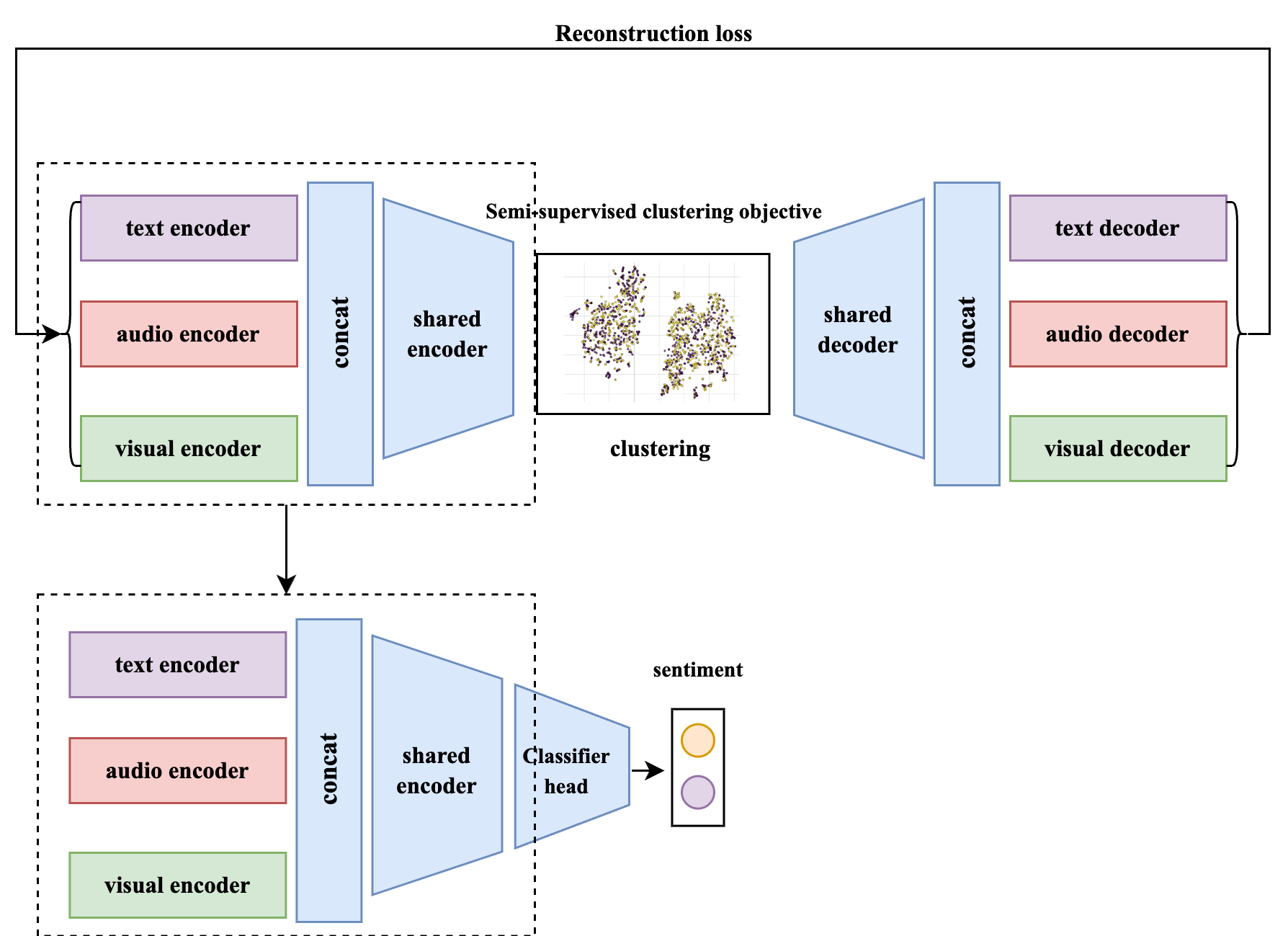}
    \caption{Network architecture schematics and the methodology overview for the semi-supervised multimodal clustering network and sentiment classification}
    \label{fig:clustering}
\end{figure}

In this phase, we enhance the baseline results by incorporating a Deep Embedded Clustering (DEC) approach, aimed at improving the representation of multimodal features through semi-supervised clustering. The DEC-based framework is designed to leverage the unlabeled data in the CMU-MOSI dataset to learn meaningful clusters within the multimodal embedding space, which are then fine-tuned for sentiment classification. Figure \ref{fig:clustering} provides an overview of the enhanced classification process using deep embedded multimodal clustering.

The DEC model is first pre-trained using a semi-supervised objective, where it attempts to cluster multimodal representations by minimizing a clustering loss. This clustering loss optimizes the network to assign each data point to the most suitable cluster based on learned features from the modalities. By utilizing this semi-supervised clustering step, we aim to capture more comprehensive, nuanced relationships across modalities, effectively embedding samples with similar sentiment patterns closer in the feature space.

The objective for the clustering network consists of the reconstruction loss for autoencoder which is typically mean-squared error and the clustering step in the original DEC approach aims to minimize the distance between the cluster assignment and a target distribution. The clustering loss is hence defined as the Kullback-Leibler (KL) divergence between the predicted distribution \( Q \) and the target distribution \( P \), which is shown in the Equation \ref{eq:clustering_loss}.

\begin{equation}
\mathcal{L}_{\text{cluster}} = \text{KL}(P \| Q) = \sum_{i} \sum_{j} P_{ij} \log \frac{P_{ij}}{Q_{ij}}
\label{eq:clustering_loss}
\end{equation}

To calculate \( Q \), the predicted soft cluster assignment, we use the Student’s t-distribution as a kernel to measure the similarity between the embedded point \( z_i \) and cluster centroid \( \mu_j \), as described in Equation \ref{eq:soft_assignment}.

\begin{equation}
Q_{ij} = \frac{\left( 1 + \| z_i - \mu_j \|^2 / \alpha \right)^{-\frac{\alpha+1}{2}}}{\sum_{j'} \left( 1 + \| z_i - \mu_{j'} \|^2 / \alpha \right)^{-\frac{\alpha+1}{2}}}
\label{eq:soft_assignment}
\end{equation}

where \( z_i \) is the latent representation of sample \( i \), \( \mu_j \) is the centroid of cluster \( j \), and \( \alpha \) is a parameter that controls the degree of freedom of the Student’s t-distribution, typically set to 1. This formulation encourages similar data points to have higher probabilities of being assigned to the same cluster. The target distribution \( P \) is then calculated in Equation \ref{eq:p}. This target distribution \( P \) reinforces high-confidence cluster assignments by assigning higher weights to points closer to cluster centroids, effectively pulling samples closer to their assigned cluster centers in the latent space. By minimizing \( \mathcal{L}_{\text{cluster}} \), we encourage the network to form compact clusters, thereby, refining the latent representation for clustering purposes.

\begin{equation}
P_{ij} = \frac{Q_{ij}^2 / \sum_{i} Q_{ij}}{\sum_{j} (Q_{ij}^2 / \sum_{i} Q_{ij})}
\label{eq:p}
\end{equation}

To complement the clustering loss, we add a reconstruction loss and a supervised term. The \textbf{reconstruction loss} in Equation \ref{eq:reconstruction_loss}, ensures latent representations \( z_i \) preserve the input structure and reconstruct \( x_i \).
\begin{equation}
\mathcal{L}_{\text{recon}} = \frac{1}{N} \sum_{i} \| x_i - \hat{x}_i \|^2
\label{eq:reconstruction_loss}
\end{equation}

We also integrate a disentanglement loss term, $\mathcal{L}_{\text{disentangle}}$, to encourage orthogonality and independence among latent representations, ensuring that different dimensions of the latent space encode distinct and uncorrelated features. This regularization helps in reducing redundancy in the latent space and facilitates better feature learning.

To compute the disentanglement loss, we first normalize the latent representations $\hat{\mathbf{z}} = \mathbf{z} - \text{mean}(\mathbf{z})$ to ensure zero-mean features. The covariance matrix $\mathbf{C}$ is then calculated as shown in Equation \ref{eq:cov}, where $N$ is the number of samples in the batch. 

\begin{equation}
C = \frac{1}{N} \hat{z}^T \hat{z}
\label{eq:cov}
\end{equation}

The disentanglement loss, $\mathcal{L}_{\text{disentangle}}$, is then defined as the sum of the squared off-diagonal elements of the covariance matrix, as shown in Equation \ref{eq:disentangle}. This penalizes correlations between different latent dimensions while preserving their variances.

\begin{equation}
\mathcal{L}_{\text{disentangle}} = \sum_{i \neq j} C_{ij}^2
\label{eq:disentangle}
\end{equation}

By minimizing $\mathcal{L}_{\text{disentangle}}$, the model learns to disentangle the latent representations, promoting independence among features and improving the interpretability and effectiveness of the learned embeddings.

We also need to account for the guiding labels. Therefore, we have utilized the typical cross-entropy loss to guide the clustering process using the labeled samples. The \textbf{supervised term} uses labeled data to guide clustering. For labeled samples $y_i$, the supervised loss is defined in Equation \ref{eq:supervised_loss}. 

\begin{equation}
\mathcal{L}_{\text{supervised}} = -\frac{1}{N_l} \sum_{i \in L} \sum_{c} y_{ic} \log \hat{y}_{ic}
\label{eq:supervised_loss}
\end{equation}

The \textbf{combined loss}, Equation \ref{eq:combined_loss} integrates all terms, weighted by $\alpha$, $\beta$ and, $\gamma$. Here $\alpha$, $\beta$ and, $\gamma$ balance the contributions of reconstruction, supervision to clustering and, the disentanglement in the latent space.

\begin{equation}
\mathcal{L}_{\text{total}} = \mathcal{L}_{\text{cluster}} + \alpha \mathcal{L}_{\text{recon}} + \beta \mathcal{L}_{\text{supervised}} + \gamma \mathcal{L}_{\text{disentangle}}
\label{eq:combined_loss}
\end{equation}

After pretraining, the DEC-based network is transferred to the sentiment classification task. The pretrained weights serve as initialization for the multimodal network, which is then fine-tuned on the labeled CMU-MOSI data. The objective in this phase is to refine the clustered representations towards a more accurate sentiment prediction model. By integrating semi-supervised clustering, this approach yields enhanced feature representations and improves classification performance.

The results of the DEC-enhanced model are compared to the baseline model, allowing us to quantify the impact of cluster-based pretraining on the final classification accuracy. This comparison provides insight into the effectiveness of incorporating clustering mechanisms for sentiment classification in multimodal settings.

\section{Experiments and Results}

To evaluate the performance of our multimodal sentiment classification framework on the CMU-MOSI dataset, we conducted a series of experiments designed to assess both the baseline model and the Deep Embedded Clustering (DEC)-enhanced model. 

\subsection{Experimental Setup}
All implementations were developed using the PyTorch framework, and experiments were conducted using an L4 GPU on Google Cloud. The experiments utilized the CMU-MOSI dataset, which was split into training and validation sets with an 80\% to 20\% ratio for both the unlabeled and labeled portions.

For each model, we monitor training and validation losses, as well as accuracy metrics, across all epochs. To measure the robustness and generalization of our approach, we report classification accuracy and F1 score. The best model checkpoint is selected based on validation accuracy and loss, and the selected model is further evaluated on the validation set for final metrics.

\subsection{Baseline Model}

In the baseline experiment, we implemented and trained the Multimodal Network model for direct sentiment classification without pretraining. This served as a foundation to assess the improvements introduced by the DEC-based semi-supervised approach. The baseline model's performance metrics, specifically accuracy and F1 score, were used as reference points to evaluate the effectiveness of DEC pretraining.

\begin{figure}[h]
    \centering
    \includegraphics[width=0.65\textwidth]{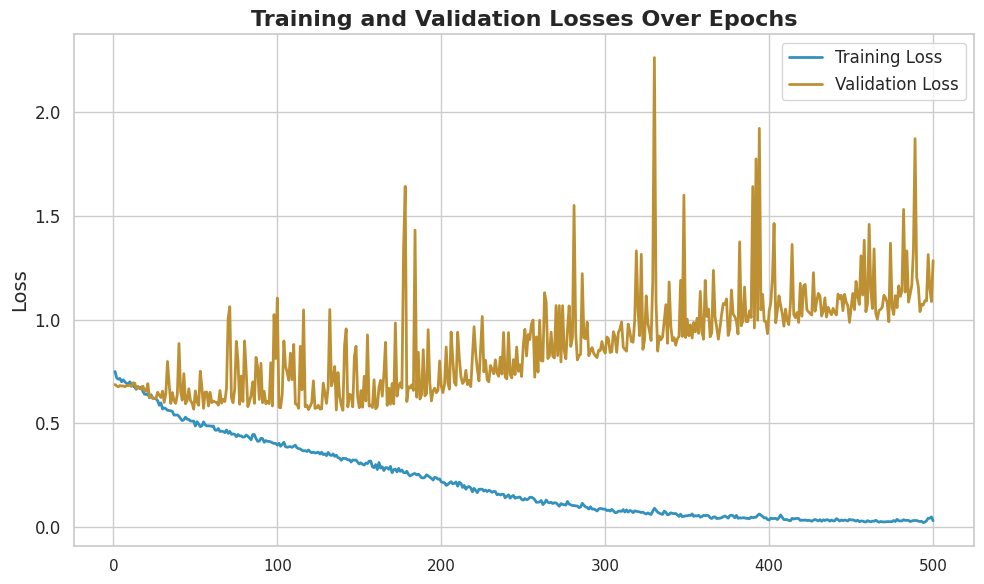}
    \caption{Loss values on training and validation set for 500 epochs of training}
    \label{fig:curves}
\end{figure}

The baseline model was trained for 500 epochs on the labeled minority of the dataset (40\%)  using the Adam optimizer with a learning rate of $4 \times 10^{-5}$ and a weight decay of $10^{-2}$ to mitigate overfitting. A batch size of 128 was used for all the experiments. After 500 epochs of training, the model achieved a training accuracy of 99.0\% and a validation accuracy of 73.14\%, along with an F1-score of 0.72. These results indicate that the accuracy values are consistent and reliable. However, the discrepancy between validation and training accuracies suggests overfitting, which could be attributed to the limited amount of labeled training data. We aim to mitigate this issue to some extent through the use of pretraining. Figure \ref{fig:curves} illustrates the progression of loss value during the training.

\subsection{DEC-enhanced Model Performance}
Training of the clustering network is performed in two stages on the entire dataset (labeled and unlabeled). We first pretrained the auto-encoder using only the reconstruction and the disentanglement terms from the Equation \ref{eq:combined_loss}. For this stage We set the $\alpha$ and $\gamma$ to 1 and 0.001 respectively and did not use the clustering term. We trained the auto-encoder for 200 epochs using the Adam optimizer with a learning rate of 5e-4. 

\begin{figure}[h]
    \centering
    \includegraphics[width=0.7\textwidth]{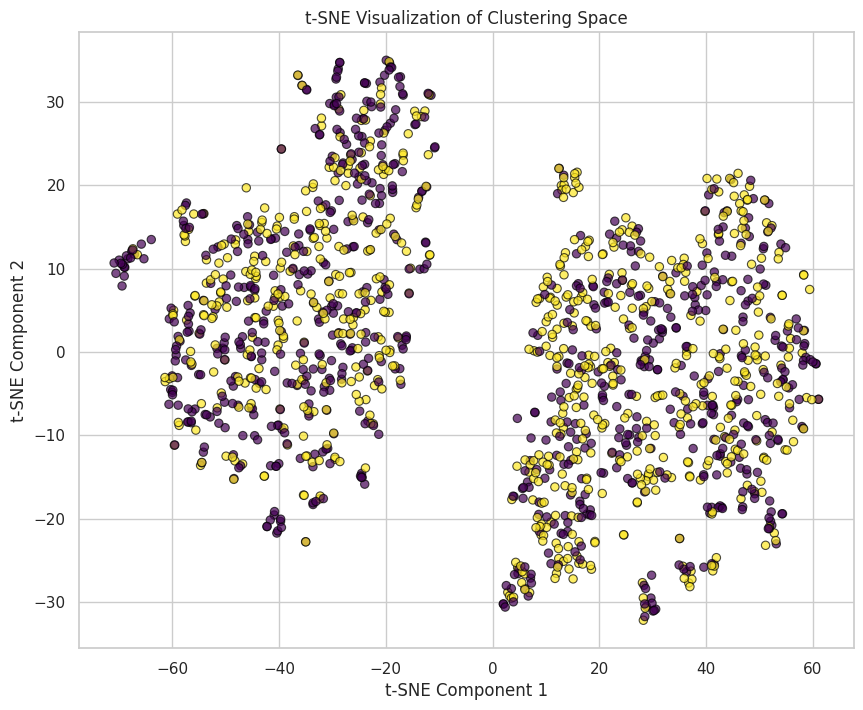}
    \caption{The t-SNE of the clustering space after the training of the network }
    \label{fig:clustering2}
\end{figure}

We then trained the autoencoder alongside the clustering network and the supervised term using the Adam optimizer with a learning rate of $1 \times 10^{-4}$ for an additional 100 epochs. We set the values of $\alpha$ and $\beta$ to 0.1 and 0.5, respectively, in Equation \ref{eq:combined_loss}, while setting $\gamma$ to zero. Figure \ref{fig:clustering2} illustrates the latent space of the clustering network, which forms clustering-friendly representations.

After training the clustering network, we leveraged the encoder portion of the pre-trained autoencoder for the classification task. This approach allowed us to harness the learned feature representations from the pretraining stage, providing an enhanced sentiment classification model compared to training the classifier from scratch.

\subsection{Results}
We trained the classifier on the labeled minority (40\%) of the dataset to simulate a scenario where labeled data is limited, while the majority of instances are unlabeled. In contrast, state-of-the-art methods have utilized the entire dataset. Despite using only 40\% of the labeled data compared to previous studies, our model's performance remains on par with them, achieving an accuracy in the range of 70\% to 80\%.

To enable a more equitable comparison of our model's capabilities, we also trained it on the full dataset after the pretraining and transfer process. This approach achieved an accuracy of 81.5\%, coming within 2\% of the best-performing model while significantly reducing the number of parameters. Table \ref{tab:comparison_results} summarizes the comparison of our results with state-of-the-art methods on the CMU-MOSI dataset. In this table we have only used accuracy for comparison since this metric was reported for all the studies.

\begin{table}[h]
    \centering
    \caption{Comparison of sentiment classification results on CMU-MOSI, including baseline, DEC-pretrained, and state-of-the-art methods, with the number of parameters for each model.}
    \label{tab:comparison_results}
    \begin{tabular}{lcc}
        \hline
        \textbf{Model} & \textbf{Accuracy (\%)} & \textbf{Param Count} \\
        \hline
        TFN \cite{gkoumas2021makes} & 74.60 & 14,707,911 \\
        DF \cite{zadeh2018multi} & 72.30 & - \\
        MARNN \cite{kim2020multi} & 84.31 & 1,350,389 \\
        MMUU-BA \cite{ghosal-etal-2018-contextual} & 82.31 & 2,424,965 \\
        \textbf{Transformer-based (40\%)} & \textbf{73.14} & 851,713 \\
        \textbf{Transformer + DEC Pretraining (40\%)} & \textbf{75.86} & 851,713 \\
        \textbf{Transformer + DEC Pretraining (Full Dataset)} & \textbf{81.50} & 851,713 \\
        \hline
    \end{tabular}
    \vspace{0.25cm}
\end{table}

\section{Discussion}

\subsection{Key Insights and Contributions}

In this study, we proposed a deep learning framework for multimodal sentiment classification and clustering using the CMU-MOSI dataset. Our methodology integrated a baseline classification network with a DEC-based semi-supervised pretraining approach to address the challenge of limited labeled data. The baseline model, trained on only 40\% of the labeled data, achieved competitive accuracy in the range of 70\% to 80\%, demonstrating its robustness under limited supervision. 

When trained on the full dataset following the pretraining and transfer process, the model achieved an accuracy of 81.5\%, within 2\% of the best-performing state-of-the-art methods. This improvement, combined with a significant reduction in parameters, underscores the efficiency of our approach.

\subsection{Limitations}

Compared to state-of-the-art methods, which leverage the full dataset for training, our framework effectively utilized both labeled and unlabeled data, achieving competitive results while requiring fewer labeled instances and reducing computational complexity. However, the discrepancy between training and validation accuracies in the baseline model suggests overfitting, likely due to the limited amount of labeled data. While pretraining mitigated this issue to some extent, further optimization of the semi-supervised learning process is necessary. Additionally, the scope of our analysis was limited to the CMU-MOSI dataset, and the generalizability of the proposed framework to other datasets and tasks remains to be validated.

\subsection{Future Work}

Future work could explore the integration of additional modalities or features to further enhance sentiment classification accuracy. Investigating alternative clustering techniques or loss functions may also improve disentanglement and clustering performance. Expanding the evaluation to other benchmark datasets and tasks would provide deeper insight into the generalizability of the proposed approach. Furthermore, incorporating domain adaptation techniques could make the model more versatile in handling cross-domain data.

\subsection{Conclusion}

In conclusion, our study demonstrates that semi-supervised clustering serves as a valid and robust pretraining objective, effectively translating to downstream classification due to the inherent similarity between clustering and classification tasks. However, utilizing the limited labeled data to guide the clustering process in the correct direction remains crucial for achieving optimal performance. 

The proposed framework achieves competitive performance with improved efficiency, making valuable contributions to the field of multimodal learning.

\bibliographystyle{unsrt}
\bibliography{reference}
\end{document}